\documentclass{Interspeech2024}

\usepackage{amsmath}
\usepackage{booktabs}
\usepackage{multirow}
\usepackage{hyperref}
\usepackage{graphicx} 
\usepackage{tabularx} 
\usepackage{pifont}
\usepackage{bbding}     
\usepackage[symbol]{footmisc}   

\newcommand\blfootnote[1]{%
  \begingroup
  \renewcommand\thefootnote{}\footnote{#1}%
  \addtocounter{footnote}{-1}%
  \endgroup
}

\interspeechcameraready

\title{Improved Factorized Neural Transducer Model For Text-only Domain Adaptation}

\name[affiliation={1}]{Junzhe}{Liu}
\name[affiliation={2}]{Jianwei}{Yu}
\name[affiliation={1}]{Xie}{Chen $^\ast$}

\address{
  $^1$ MoE Key Lab of Artificial Intelligence, AI Institute, X-LANCE Lab, Department of Computer Science and Engineering,  Shanghai Jiao Tong University, Shanghai, China\\
  $^2$ Tencent AI Lab, Shenzhen, China
}
\email{junzheliu@sjtu.edu.cn, tomasyu@tencent.com, chenxie@sjtu.edu.cn}

\keywords{neural Transducer, text-only domain adaptation, end-to-end speech recognition, language model}

\begin{document}
\maketitle

\begin{abstract}
    
Adapting End-to-End ASR models to out-of-domain datasets with text data is challenging. Factorized neural Transducer (FNT) aims to address this issue by introducing a separate vocabulary decoder to predict the vocabulary. Nonetheless, this approach has limitations in fusing acoustic and language information seamlessly. Moreover, a degradation in word error rate (WER) on the general test sets was also observed, leading to doubts about its overall performance.
In response to this challenge, we present the improved factorized neural Transducer (IFNT) model structure designed to comprehensively integrate acoustic and language information while enabling effective text adaptation.
We assess the performance of our proposed method on English and Mandarin datasets. The results indicate that IFNT not only surpasses the neural Transducer and FNT in baseline performance in both scenarios but also exhibits superior adaptation ability compared to FNT. On source domains, IFNT demonstrated statistically significant accuracy improvements, achieving a relative enhancement of $1.2\%$ to $2.8\%$ in baseline accuracy compared to the neural Transducer.
On out-of-domain datasets, IFNT shows relative WER(CER) improvements of up to $30.2\%$ over the standard neural Transducer with shallow fusion, and relative WER(CER) reductions ranging from $1.1\%$ to $2.8\%$ on test sets compared to the FNT model.\blfootnote{$\ast$ Corresponding author}
\end{abstract}


\section{Introduction}
\label{sec:intro}

In recent years, end-to-end (E2E) \cite{li2022recent} based models have gained great interest in automatic speech recognition (ASR) systems. Compared to traditional hybrid systems, E2E systems such as connectionist temporal classification (CTC) \cite{ctc}, attention-based encoder-decoder (AED) \cite{las}, and neural Transducer (NT)  \cite{sequence} predict word sequences using a single neural network. When there is a mismatch between the trained domain and the test domain, a significant degradation in accuracy is observed. Conventional domain adaptation methods   \cite{Bell_2021, deng2023adaptable} typically rely on speech-text pairs from the target domain. However, collecting a large amount of speech-text matching data from the target domain is difficult, while obtaining text-only data is relatively easier. As a result, text-only adaptive methods have been widely proposed and studied  \cite{pylkkonen2021fast, Choudhury2022}. Since E2E systems are jointly optimized, there is no separate component that solely performs as a language model (LM), making it challenging to directly apply common LM adaptation methods.

One feasible solution is to fine-tune the E2E model using the synthesized audio-transcript pairs generated by a text-to-speech (TTS) model  \cite{sim2019personalization, zheng2021using, deng2021improving}, but this approach is computationally expensive. Another common practice is LM fusion  \cite{cabrera2021language, 8639038, levit2023external, li2023prompting}, such as shallow fusion  \cite{kannan2017analysis}, deep fusion  \cite{gulcehre2015using}, and cold fusion  \cite{coldfusion}. Among them, the most widely used is shallow fusion, which combines the E2E model score and the external LM score in the log-linear domain during beam search. Methods like density ratio  \cite{mcdermott2020density} also work similarly. Internal language model estimation  \cite{ilm, meng2021internal} was also proposed recently, it calculates the interpolated log-likelihood score based on the maximum scores from the internal LM and the external LM respectively during decoding. However, LM fusion involves interpolation weights that are task-dependent and require tuning, making the performance sensitive to the weight selection. 

There have been increasing research efforts  \cite{variani2020hybrid, factorizedAED, mhat} to modify the structure of neural Transducers. Factorized neural Transducer (FNT)   \cite{zhao2023fast, chen2021Factorized,levit2023external, gong2023longfnt,le2023factorized} addresses this issue by introducing a standalone LM for vocabulary prediction, enabling the direct application of conventional LM adaptation methods. While the results have shown promising adaptation abilities, minor accuracy degradation has also been observed on the general test set compared to the standard neural Transducer baseline, raising doubts about their overall performance. 

Building upon this work, we propose the improved factorized neural Transducer (IFNT) model that can better combine acoustic information with language information, and improve its baseline accuracy both before and after adapting to text data. Our proposed model incorporates an internal LM alongside the standard neural Transducer model, with the LM posterior probability directly integrated into the vocabulary prediction. Utilizing a standalone LM in our model facilitates the application of various LM adaptation methods for fast text-only domain adaptation, similar to the hybrid system. We validate the proposed method in both English and Mandarin datasets, both in-domain 
and out-of-domain scenarios 
results demonstrate the superior performance of our proposed model over the FNT and standard neural Transducer shallow fusion methods.

\section{Standard and Factorized Neural Transducer Models}
\label{sec:related}

\subsection{Standard neural Transducer}


ASR systems predicts a conditional distribution over blank-augmented token sequences $\hat{\mathbf{Y}}=\left\{\hat{y}_1, \ldots, \hat{y}_{T+U}\right\}$, where $T, U$ are acoustic and label sequence lengths, $\hat{y}_i \in \mathcal{V} \cup\phi$, and $\mathcal{V}, \phi$ denotes vocabulary and blank respectively. The standard neural Transducer model can be divided into three parts: an acoustic encoder, which takes the acoustic feature $\boldsymbol{x}_1^t$ and generates the acoustic representation $\bf{f_t}$; a label decoder, which consumes the history of the previously predicted label sequence $\boldsymbol{y}_1^u$ and computes the label representation $\bf g_u$; and a joint network, which takes both representations and combines them to compute the probability distribution over $\mathcal{V} \cup\phi$:

\begin{equation}
P\left(\hat{y}_{t+1} \mid \mathbf{x}_1^t, \mathbf{y}_1^u\right)=\operatorname{softmax}\left(\sigma(\bf{f_t} + \bf g_u)\right)
\end{equation}

$\sigma$ denotes some non-linear activation function, e.g. relu. In Figure 1(a), the part enclosed by the gray dashed box is the joint network. It first projects the outputs of the encoder and decoder to a joint dimension (represented as $D$ in the figure), then adds the two together, and finally maps the output to a $V+1$ dimension corresponding to the vocabulary plus the blank token. The objective function of the neural Transducer is to minimize the negative log probability over all possible alignments, which could be written as:

\begin{equation}
\mathcal{J}_t=-\log P\left(\mathbf{Y} \mid \mathbf{x}\right)=-\log \sum_{\alpha \in \beta^{-1}(\mathbf{y})} P(\alpha \mid \mathbf{x})
\end{equation}

where $\beta$ is the function to convert the alignment $\alpha$ to label sequence $\mathbf{Y}$ by removing the blank $\phi$.

\subsection{LM shallow fusion}

In shallow fusion, an LM trained on target domain training text is integrated with the E2E model during inference to optimize a log-linear interpolation between the E2E and LM probabilities. The optimal token sequence $\mathbf{Y}$ is obtained via beam search:

\begin{equation}
\mathbf{Y}=\underset{\mathbf{Y}}{\arg \max }\left[\log P\left(\mathbf{Y}\mid\mathbf{X};\theta_{\mathrm{E}2\mathrm{E}}^{\mathrm{S}}\right)+\lambda_T \log P\left(\mathbf{Y} ; \theta_{\mathrm{LM}}^{\mathrm{T}}\right)\right]
\end{equation}

where $P\left(\mathbf{Y}; \theta_{\mathrm{LM}}^{\mathrm{T}}\right)$ is the posterior probability given by the external LM, and $\lambda_T$ is a hyper-parameter for tuning.

\subsection{Factorized Neural Transducer}

Although the neural Transducer's structure includes a label decoder, it is not entirely equivalent to a language model. Because the output of the decoder is a high dimensional representation of the tokens, rather than a posterior distribution; Moreover, typical LM only predicts the vocabulary tokens $\mathcal{V}$, while the joint network also has to predict the blank token $\phi$.

Recognizing this distinction, the factorized neural Transducer (FNT) adopts a structure consisting of two separate decoders. As shown in Fig 1(b), the original joint network portion (enclosed in the gray box) remains the same as the standard neural Transducer structure, with the exception that the projection layer maps the output to a dimension of $1$, predicting only the blank token. Consequently, this decoder is referred to as the blank decoder.
In the vocabulary section, FNT introduces a separate language model component into the model, generating a probability distribution over $\mathcal{V}$. The encoder output is projected to a dimension of $V$ (indicated by the yellow projection layer). Acoustic and label information are then combined at the logit level to predict vocabulary tokens. The logit for the blank token and the vocabulary logits are concatenated to compute the Transducer loss. The total training loss of FNT is expressed as:

\begin{equation}
\mathcal{J}_f=\mathcal{J}_t-\lambda_f \log P_{LM}\left(\mathbf{Y}\right)
\end{equation}

where the first term is the Transducer loss, and the second term is the language model loss with cross-entropy. $\lambda_f$ is a hyper-parameter for LM tuning. Since the vocabulary decoder works as a standalone language model, we could use the target domain's text data to fine-tune this part directly.

\section{Improved factorized neural Transducer}
\label{sec:improved}

\begin{figure*}[htb]
\centering
\includegraphics[width=1.0\linewidth]{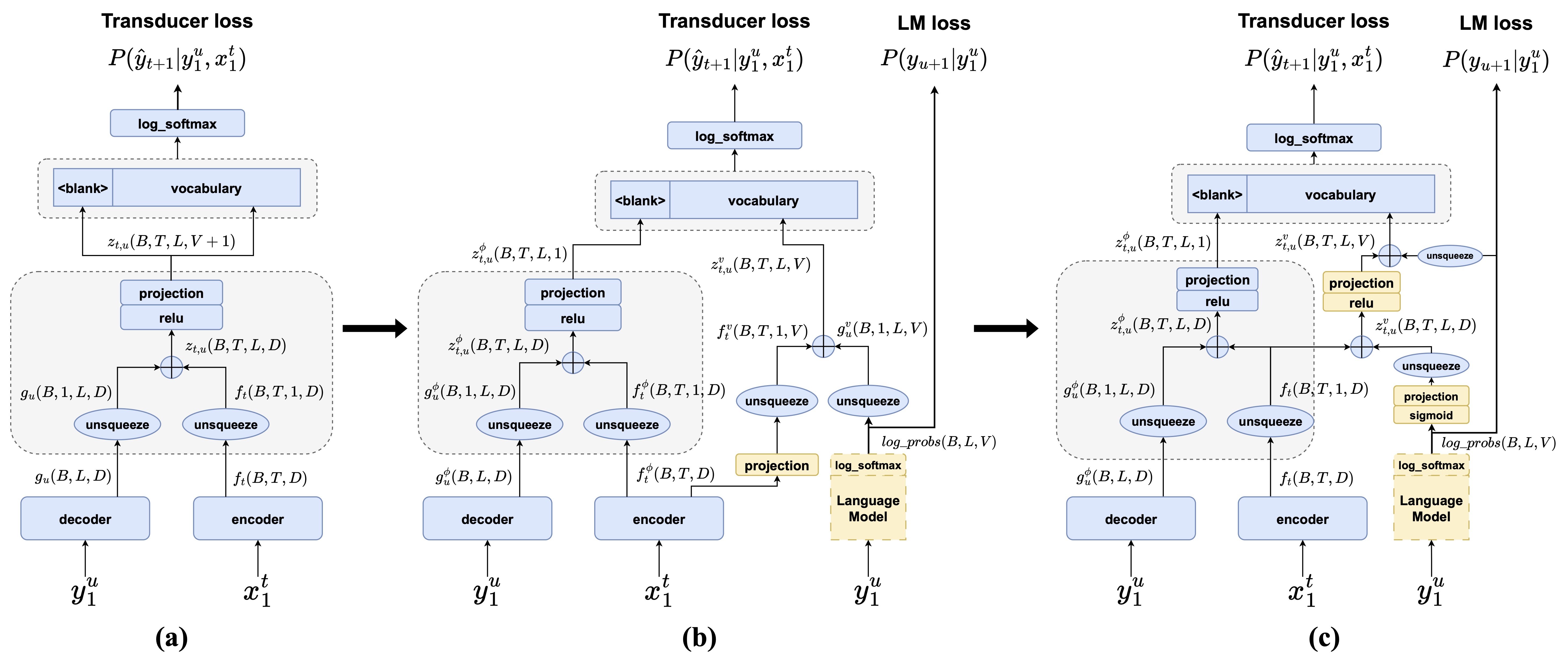} 
\caption{The illustration of three model structures: (a) standard neural Transducer (NT). (b) factorized neural Transducer (FNT). (c) proposed improved factorized neural Transducer (IFNT).}
\label{improved}
\end{figure*}

Despite the good adaptability of FNT, an accuracy degradation on general test sets was also observed. To narrow this gap, we propose the IFNT model with enhanced performance and adaptability. We noted that FNT employs a different approach in combining acoustic and label information compared to the standard neural Transducer, in which the encoder and decoder output are fused in the joint dimension space ($D$-dimension) before mapping to the vocabulary size. In contrast, FNT directly adds both information together in the vocabulary space ($V$-dimension) on a logit basis. Inspired by this difference, we revert to the standard neural Transducer style of integrating these two types of information in our proposed IFNT model. 
An illustration of our IFNT model is presented in Fig 1(c).

We introduced several major modifications in IFNT: 
Firstly, we apply a sigmoid layer to the output of the vocabulary decoder. Our goal is to constrain the distribution of text vectors in the feature space, thereby facilitating faster convergence during adaptation.
Secondly, LM output is projected to the joint dimension $D$ and then fused with the encoder output, following the standard neural Transducer approach.
Furthermore, we directly incorporate the LM posterior probability over $\mathcal{V}$ into the final output of the vocabulary component. 
Including the probability distribution in the final layer helps ensure the preservation and effective integration of essential linguistic information with acoustic features, resulting in improved recognition accuracy. 
The total training loss is the same as FNT.


Compared to deep fusion  \cite{gulcehre2015using} and cold fusion  \cite{coldfusion}, our proposed IFNT model stands out in that the E2E model and LM component are jointly trained from scratch, while deep fusion and cold fusion rely on pre-trained LMs for integration. Moreover, the most significant distinction lies in the direct incorporation of the LM posterior probability into the final output, without which the IFNT model would lose its adaptation ability.
We have also explored the use of a pre-trained language model for LM initialization in IFNT. However, experiments revealed that training the whole model from scratch yields more stable results. 
This may be because joint training allows the LM to synchronize more effectively with the acoustic encoder.


\section{Experiments}
\label{sec:experiments}

\subsection{Datasets}

We evaluate our model on both English and Mandarin datasets. In the English context, we utilize the GigaSpeech-M  \cite{chen2021gigaspeech} subset as our training set, comprising 1,000 hours of speech data. For text-only domain adaptation, we have chosen three target domain datasets: EuroParl \cite{europarl}, TED-LIUM \cite{tedlium}, and a Medical \cite{medical} dataset. In the Mandarin scenario, we employ the Wenetspeech \cite{zhang2022wenetspeech} train-M subset, which encompasses 1,000 hours of speech data, as our training set. Additionally, for out-of-domain adaptation, we incorporate four datasets: Thchs-30 \cite{wang2015thchs}, Aishell-1  \cite{bu2017aishell}, Aishell-2 (iOS)  \cite{du2018aishell}, and Aishell-4 (L subset) \cite{fu2021aishell}. Detailed dataset statistics can be found in table 1.



\begin{table}[h]
\centering
\caption{Statistics of the source domain dataset (GigaSpeech, Wenetspeech) and target domain datasets including their durations in hours (h) and number of training text sentences.\\}
\begin{tabular}{l|rrrr}
\toprule
\textbf{Dataset} & \textbf{\#sentences} & \textbf{Train} & \textbf{Dev} & \textbf{Test} \\
\midrule
\textbf{GigaSpeech-M} & 909,401 & 1000.0 & 12.0 & 40.0 \\
EuroParl & 32,584 & 77.2 & 3.1 & 2.9 \\
TED-LIUM & 268,263 & 346.2 & 3.7 & 3.8 \\
Medical & 20,107 & 35.1 & 4.9 & 5.2 \\
\midrule
\textbf{Wenetspeech-M} & 1,514,500 & 1000.0 & 20.0 & 38.0 \\
Thchs-30 & 10,001 & 27.1 & - & 6.2 \\
Aishell-1 & 120,099 & 150.0 & 10.0 & 5.0 \\
Aishell-2 (iOS) & 1,009,224 & 1000.0 & 2.5 & 5.1 \\
Aishell-4 (L) & 12,494 & 16.2 & - & 12.7 \\
\bottomrule
\end{tabular}
\label{tab:table1}
\end{table}

\begin{table}[h]
\renewcommand{\arraystretch}{1.1}
\centering
\caption{Comparison of ASR accuracy on GigaSpeech (WER $\%$) and Wenetspeech (CER $\%$) between standard neural Transducer, FNT, and IFNT. Model parameters are all 106M.\\}
\begin{tabular}{l|cc|cc}
\toprule
\multirow{2}{*}{\textbf{Model}} & \multicolumn{2}{c|}{\textbf{GigaSpeech}} & \multicolumn{2}{c}{\textbf{Wenetspeech}} \\

 & \textbf{dev} & \textbf{test} & \textbf{dev} & \textbf{test\_net}\\

\midrule

\ding{192} Transducer & 13.96 & 13.99 & 10.80 & 11.25\\
\ding{193} FNT & 14.11 & 14.19 & 11.29 & 11.76\\
\ding{194} IFNT (proposed) & \textbf{13.67} & \textbf{13.84} & \textbf{10.77} & \textbf{11.23} \\

\bottomrule
\end{tabular}
\label{tab:table4}
\end{table}

\begin{table*}[h]
\renewcommand{\arraystretch}{1.1}
\centering
\caption{Comparison of Perplexity (PPL) on the dev set and ASR accuracy among the standard neural Transducer, FNT, and IFNT model before and after adaptation. Upper table: Results for English datasets (WER $\%$). Lower table: Results for Mandarin datasets (CER $\%$). The number of parameters for all three models is equally set to 106M, with the external language model used for shallow fusion comprising 4M parameters.\\}
\begin{tabular}{l|c|ccc|ccc|ccc|c}
\toprule
\multirow{2}{*}{\textbf{Model}} & \textbf{text} & \multicolumn{3}{c|}{\textbf{EuroParl}} &  \multicolumn{3}{c|}{\textbf{TED-LIUM}} & \multicolumn{3}{c|}{\textbf{Medical}} & \textbf{Avg. }\\ 

 & \textbf{adaptation} & \textbf{PPL} & \textbf{dev} & \textbf{test}  & \textbf{PPL} & \textbf{dev} & \textbf{test}  & \textbf{PPL} & \textbf{dev} & \textbf{test} & \textbf{WER}\\
 
\midrule

\ding{192} neural Transducer & \XSolidBrush &  - & 20.22 & 20.81 & - & 11.17 & 11.90 & - & 29.12 & 28.47 & 20.28\\

\ding{193} FNT & \XSolidBrush & 107.31 & 20.53 & 21.21 & 120.13 & 11.12 & 11.91 & 173.12 & 29.39 & 28.64 & 20.47\\

\ding{194} IFNT (proposed) & \XSolidBrush & 102.91 & \textbf{20.21} & \textbf{20.69} & 123.30 & \textbf{11.11} & \textbf{11.88} & 158.33 & \textbf{28.63} & \textbf{28.20} & \textbf{20.12}\\

\midrule

\ding{192} + shallow fusion & \CheckmarkBold  & 55.89 & 19.86 & 20.19 & 67.83 & 10.86 & 11.71 & 18.46 & 28.09 & 27.10 & 19.64\\

\ding{193} + text-only finetune & \CheckmarkBold  & 37.65 & 18.15 & 18.80 & 61.03 & 8.55 & 8.38 & 12.06 & 22.16 & 21.51 & 16.26\\

\ding{194} + text-only finetune & \CheckmarkBold & 39.84 & \textbf{18.09} & \textbf{18.27} & 60.44 & \textbf{8.21} & \textbf{8.17} & 11.61 & \textbf{21.91} & \textbf{21.50} &  \textbf{16.02}\\

\bottomrule
\end{tabular}
\label{tab:table2}
\end{table*}

\begin{table*}[h]
\renewcommand{\arraystretch}{1.1}
\centering
\begin{tabular}{l|c|cc|ccc|ccc|cc|c}
\toprule
\multirow{2}{*}{\textbf{Model}} & \textbf{text} & \multicolumn{2}{c|}{\textbf{Thchs-30}} &  \multicolumn{3}{c|}{\textbf{Aishell-1}} & \multicolumn{3}{c|}{\textbf{Aishell-2}} & \multicolumn{2}{c|}{\textbf{Aishell-4 }}& \textbf{Avg. } \\ 

 & \textbf{adaptation} & \textbf{PPL} & \textbf{test}  & \textbf{PPL} & \textbf{dev} & \textbf{test}  & \textbf{PPL} & \textbf{dev} & \textbf{test} & \textbf{PPL} & \textbf{test} & \textbf{CER}\\
 
\midrule

\ding{192} neural Transducer & \XSolidBrush & - & 10.19 & - & \textbf{5.57} & 6.60 & - & \textbf{6.37} & \textbf{6.39} & - & 32.38 & 11.25\\

\ding{193} FNT & \XSolidBrush & 289.31 & 10.51 & 73.25 & 5.97 & 6.67 & 68.93 & 6.77 & 6.63 & 81.24 & 31.59 & 11.36\\

\ding{194} IFNT (proposed) & \XSolidBrush & 275.61 & \textbf{10.07} & 73.11 & 5.73 & \textbf{6.49} & 69.90 & 6.64 & 6.49 & 77.91 & \textbf{31.29} & \textbf{11.12}\\

\midrule

\ding{192} + shallow fusion & \CheckmarkBold & 183.11 & 10.01 & 57.81 & 5.27 & 6.14 & 58.14 & 6.27 & 6.24 & 67.01 & 30.97 & 10.82 \\

\ding{193} + text-only finetune & \CheckmarkBold & 214.16 &  9.99 & 43.80 & 4.97 & 5.57 & 47.28 & 6.43 & 6.24 & 39.09 & 29.91 & 10.52\\

\ding{194} + text-only finetune & \CheckmarkBold & 208.11 & \textbf{9.84} & 46.89 & \textbf{4.90} & \textbf{5.54} & 50.44 & \textbf{6.27} & \textbf{6.17} & 38.64 & \textbf{29.88} & \textbf{10.43} \\

\bottomrule
\end{tabular}
\label{tab:table3}
\end{table*}

\subsection{Experiment setups}

We build 80-dimensional mel filterbank features with global-level cepstral mean and variance normalization for acoustic feature extraction. Regarding the model structure, the encoder consists of 12 Conformer  \cite{gulati2020conformer} layers. The inner size of the feed-forward layer is 2,048, and the attention dimension is 512 with 8 heads. To maintain a roughly equal number of parameters across three models, we keep the encoder part consistent while modifying the decoder's structural parameters, and the final number of parameters of the models remains around 106M. The hyperparameters are tuned for the best accuracy, and the final choices are $\lambda_T=0.1$ and $\lambda_f=0.1$. Our experiments are implemented using the fairseq \cite{ott2019fairseq} framework. During adaptation, factorized Transducer models are fine-tuned using the target domain's training text and then evaluated on the test sets.
Our training utilizes fp32 precision and the Adam optimizer.
All models are trained under the same training configuration. We apply a model averaging over 5 checkpoints during inference, and beam search with a beam size of 5 is used.

\subsection{In-domain evaluation}

Table 2 illustrates the accuracy of the three models on the source domain test sets.
IFNT exhibits superior baseline accuracy over both Transducer and FNT across English and Mandarin datasets, achieving noteworthy relative WER(CER) reductions ranging from $2.5\%$ to $4.6\%$ compared to FNT, and $1.2\%$ to $2.8\%$ compared to the neural Transducer.


\subsection{Out-of-domain evaluation}

In the out-of-domain scenarios, we evaluate the accuracy of the models both before and after adaptation to training texts. For the standard neural Transducer, we incorporate an external language model trained on the training text with a parameter of $4M$ for shallow fusion. Results are presented in Table 3.

In the English scenario, IFNT exhibits significant improvements across three datasets. Prior to text-only adaptation, IFNT outperforms FNT in baseline results and surpasses the neural Transducer, achieving up to $2.6\%$ relative WER reductions. After fine-tuning with target domain training text, IFNT demonstrates substantial relative WER reductions of $9.5\%$, $30.2\%$, and $20.7\%$ compared to the neural Transducer with shallow fusion on the respective test sets. In comparison to FNT, IFNT displays enhanced adaptation capabilities, resulting in relative WER reductions of $2.8\%$, $2.5\%$, and $1.1\%$ respectively. 

For the Mandarin datasets, IFNT has surpassed the baseline accuracy of FNT entirely and demonstrated better results over the conventional neural Transducer on three datasets. Following text adaptation, IFNT achieves the lowest CER results, exhibiting a relative CER reduction of $1.1\%$ to $9.8\%$ compared to the neural Transducer with shallow fusion. Furthermore, in comparison to FNT, IFNT demonstrates a relative CER improvement of up to $2.5\%$ after completing text adaptation.


\subsection{Ablation study and analysis}

We tried to re-model the decoder network as a language model thereby eliminating the need for a separate third component. However, this alteration did not result in any enhancement in the overall accuracy, nor did it succeed in adapting to various domains. The findings clearly show that a distinct language model component is essential for the desired adaptability. 

We then conducted ablation experiments to validate the efficacy of the modifications introduced in IFNT by removing the log\_probs directly injected into the final projection layer, which is denoted as \textit{IFNT w/ lprobs} and index 4. The most significant distinction between this model and FNT lies in the structure of vocabulary prediction. The baseline results are presented in Table 4, indicating that the structural modifications indeed led to improvements in baseline accuracy compared to FNT.

To assess its adaptation ability, we chose EuroParl and Aishell-1 as out-of-domain test sets. Experimental results reveal a substantial decline in IFNT's adaptation capability upon the removal of log\_probs, emphasizing the crucial role of incorporating log\_probs for maintaining adaptation capability.

\begin{table}[h]
\centering
\caption{Ablation study:  \ding{195} \textit{IFNT w/ lprobs} denotes the IFNT model after removing the log\_probs. Upper table: Baseline results on source domain datasets. Lower table: Adaptation results on target domain datasets. \\}
\begin{tabular}{l|cc|cc}
\toprule
\multirow{2}{*}{\textbf{Model}} & \multicolumn{2}{c|}{\textbf{GigaSpeech}} & \multicolumn{2}{c}{\textbf{Wenetspeech}} \\

 & \textbf{dev} & \textbf{test} & \textbf{dev} & \textbf{test\_net}\\

\midrule

\ding{193} FNT & 14.11 & 14.19 & 11.29 & 11.76\\

\ding{195} IFNT w/ lprobs & \textbf{13.48} & \textbf{13.49} & 11.26 & 11.75\\

 \ding{194} IFNT (proposed) & 13.67 & 13.84 & \textbf{10.77} & \textbf{11.23}\\

\bottomrule
\end{tabular}
\label{tab:table4}
\end{table}

\begin{table}[h]
\centering
\begin{tabular}{l|ccc|ccc}
\toprule
\multirow{2}{*}{\textbf{Model}} & \multicolumn{3}{c|}{\textbf{EuroParl}} & \multicolumn{3}{c}{\textbf{Aishell-1}} \\

 & \textbf{PPL} & \textbf{dev} & \textbf{test} & \textbf{PPL} & \textbf{dev}& \textbf{test} \\

\midrule

\ding{195} & 97.89 & 20.16 & 20.25 & 87.23 & 6.01 & 6.93 \\

\ding{195} + adapt & 47.66 & 20.03 & 20.21 & 56.84 & 6.01 & 6.90 \\
\bottomrule
\end{tabular}
\label{tab:table4}
\end{table}

In summary, our modifications in IFNT prove beneficial to the model's baseline and adaptation results, aligning with the findings of \cite{graves2013speech} that earlier fusion of acoustic and linguistic information leads to better performance in Transducer models.

\subsection{Limitations}

Despite our proposed IFNT exhibiting impressive baseline and adaptation performances, it still presents certain limitations: 1) It does not leverage audio data for adaptation, and 2) It still remains confined within the Transducer framework structure. 
Future exploration could consider re-designing the blank token $\phi$ for further improvements.

\section{Conclusion}
\label{sec:conclusion}

In this work, we proposed the IFNT model. By redesigning the model structure of the FNT, we addressed the drawback of FNT experiencing a drop in baseline accuracy compared to the neural Transducer, while also enhancing its capability for text-only domain adaptation. On both Mandarin and English datasets, IFNT demonstrated statistically significant accuracy improvements, achieving a relative enhancement of $1.2\%$ to $2.8\%$ in baseline accuracy compared to the neural Transducer. On out-of-domain datasets, its adaptation capability demonstrated a relative reduction of up to $2.8\%$ compared to FNT and up to $30.2\%$ compared to the neural Transducer.


\vfill\pagebreak





\section{Acknowledgements}

This work was supported by the National Natural Science Foundation of China (No.62206171 and No.U23B2018), Shanghai Municipal Science and Technology Major Project under Grant 2021SHZDZX0102, and the International Cooperation Project of PCL.

\bibliographystyle{IEEEtran}
\bibliography{mybib}

\end{document}